\documentclass[letterpaper]{article}

\usepackage[top=1in, bottom=1.25in, left=1.25in, right=1.25in]{geometry}
\usepackage[sort]{cite}
\usepackage{times}
\usepackage{url}
\usepackage{latexsym}

\usepackage{amsmath}
\usepackage{graphicx}
\usepackage{multirow}
\usepackage{mdframed}

\title{Incremental Learning for Fully Unsupervised Word Segmentation Using
Penalized Likelihood and Model Selection}

\author{
  Ruey-Cheng Chen\footnote{National Taiwan University.  email: {\tt rueycheng@ntu.edu.tw}}
}

\date{Sep 1, 2014}

\begin{document}
\maketitle

\begin{abstract}

  We present a novel incremental learning approach for unsupervised word
  segmentation that combines features from probabilistic modeling and model
  selection.  This includes super-additive penalties for addressing the
  cognitive burden imposed by long word formation, and new model selection
  criteria based on higher-order generative assumptions.  Our approach is fully
  unsupervised; it relies on a small number of parameters that permits flexible
  modeling and a mechanism that automatically learns parameters from the data.
  Through experimentation, we show that this intricate design has led to
  top-tier performance in both phonemic and orthographic word segmentation.

\end{abstract}

\section{Introduction}

Information criteria are meta-heuristics that measure the quality of
statistical fits.  Well-known examples such as Akaike information criterion
\cite{akaike1974new} and minimum description length \cite{rissanen1978modeling}
were developed in the context of model selection.  These powerful tools deal
directly with statistical fits regardless of how these models are inferred,
thereby allowing one to efficiently test a large amount of hypotheses of
different distributional assumptions.  

Despite being around for many decades, this level of optimization was scarce in
computational linguistics.  One reason is that most solutions developed in the
past for linguistic problems rest on fairly simple distributional assumptions,
and few needs to go beyond conventional techniques for the inference.  But as
many recent methods are beginning to embrace sophisticated structural
assumptions such as context or syntax, the complexity of inference can be
overwhelmingly large.  One such troubling case is the nonparametric Bayesian
approach for unsupervised word segmentation.  Nonparametric Bayesian modeling
\cite{goldwater2009bayesian,johnson2009improving} is commonly deemed as state
of the art for unsupervised word segmentation, but the computation overhead has
largely impeded its application on orthographic data, such as texts written in
Chinese, Japanese, ancient languages, or even those with little linguistic
resources.

This need for efficiently processing large text body has inspired some focused
deployment of minimum description length (MDL) into the area
\cite{zhikov2010efficient,hewlett2011fully,chen2012regularized}.  Troubled by
the lack of efficient search procedure for MDL, these methods use specialized
algorithms to explore the hypothesis space.  The search algorithm is usually
light on execution and takes one or more input parameters that serve as knobs
for biasing search direction.  It is therefore easy to induce many ``search
models'' at the same time by just changing the parameters, and then combine
these outputs altogether using information criteria such as MDL.  This approach
has proven to be efficient and scales well to large benchmarks in orthographic
word segmentation.  But as these search algorithms lean more towards using
simple heuristics and hardly take structural assumptions into account,
MDL-based methods are sometimes criticized for having inferior performance in
the modeling of language.

% have relied on \emph{incremental learning}, a greedy algorithm that checks only
% nearby configurations for reducing description length or some other criteria,
% to explore the solution space.  And indeed incremental learning still needs to
% rest on specific notions of nearness, but this is a lot easier than developing
% a set of efficient, non-probabilistic characterizations of the search space.
% Generally, these incremental learning methods rely on simple algorithms to
% induce as many feasible outputs as possible, and then use MDL to find the best
% segmentation result.  This new approach has proven to be very efficient, and it
% has served as a plausible alternative to more principled Bayesian approaches on
% large datasets.

% We develop a simplistic
% framework based on a basic search procedure, using a maximum likelihood
% estimate as the objective.  However, incremental learning using maximum
% likelihood alone may get stuck in local maxima very quickly and result in poor
% segmentation accuracy.  

In this paper, we seek to bridge this gap between more sophisticated
probabilistic models and more efficient MDL-based methods.  We develop a fully
unsupervised, simplistic approach called \emph{incremental learning} that
combines the best of both worlds:  We use a model selection framework for
efficiently testing hypotheses, but for better language modeling we use a
probabilistic search procedure that optimizes over the penalized likelihood.
The incremental learning approach is so named because the search procedure is a
greedy algorithm that checks only nearby configurations.  Each search instance
maintains an assumption about the language to be modeled; in each iteration,
the instance would incrementally update its hypothesis based on the boundary
placement decision it has made so far.  Also it is known that using
over-simplistic generative assumptions can easily lead to undersegmentation.
To fix this problem, we propose using a new length-based penalty term as in
Bayesian modeling \cite{goldwater2009bayesian}.  This penalty term is developed
to approximate the cognitive burden (the number of meanings, in a sense)
imposed by long word formation.  Thus it is introduced into our method as a
super-additive function, which penalizes longer words harder than shorter ones.
We compared this incremental learning algorithm with many reference methods on
both phonemic and orthographic word segmentation tasks.  The result shows that
the performance of incremental learning surpassed existing MDL-based algorithms
and compared well to the best records on both benchmarks.  

The rest of the paper is organized as follows.
Section~\ref{s:unsupervised-word-segmentation} briefly reviews the development
of unsupervised word segmentation and some of the recent advances.  In
Section~\ref{s:incremental-learning}, we describe the proposed method, model
selection criteria, and some other design features.  Section~\ref{s:experiment}
covers our test result on two standard benchmarks for phonemic and orthographic
word segmentation.  The analysis is expanded later in
Section~\ref{s:post-hoc-analysis} to look at the correlation between F-score
and information criteria.  We give out concluding remarks in
Section~\ref{s:conclusion}.

% Issue (i) is commonly understoood as a lack of efficient
% procedure to explore the search space, even though this may also be an
% advantage (for not able to throwing dice and thus staying efficient).  
% Some
% hard problems live in a peculiar search space that cannot be easily optimized
% over but may have a (sophisticated) probabilistic solution.  MDL on this type
% of problem usually ends up using greedy search that checks only the ``nearby''
% hypotheses and incrementally reduces description length.  

% This
% reliability issue has been confirmed for unsupervised word segmentation in a
% recent study on orthographic Chinese \cite{magistry2013mdl}.  

% It therefore lead us to consider a second plausible
% fix, which is to simply replace MDL with other information criterion, although
% the effect of such replacement on solution quality remains unclear to many of us.

% Take unsupervised word segmentation.  A common way to cope with the first
% problem is to use an algorithm that explores the search space to collect a set
% of candidate segmentations.  Each of these segmentation in general will
% correspond to a hypothesis in the defined search space.  Later these hypotheses
% are judged using model selection criterion such as MDL.

\section{Unsupervised Word Segmentation}\label{s:unsupervised-word-segmentation}

Unsupervised word segmentation has been a focus in natural language processing
for certain East-Asian languages such as Japanese or Chinese.  Some early
traces of unsupervised word segmentation dates back to 1950s in linguistic
studies.  Harris~\cite{harris1955phoneme} studied morpheme segmentation and
suggested that low transition probability between two phonemes may indicate a
potential morpheme boundary.  This idea has been a major influence on several
approaches in orthographic Chinese word segmentation, such as accessor variety
\cite{feng2004accessor} and branching entropy
\cite{tanaka-ishii2005entropy,jin2006unsupervised}.  Besides the focus on word
boundaries, some effort has been put into measurement of the quality of
word-like strings.  Sproat and Shih~\cite{sproat1990statistical} used pointwise mutual
information to measure the association strength of a string in Chinese texts.
Kit and Wilks~\cite{kit1999unsupervised} proposed using the amount of information gained
by modeling subword units as a single word to measure word quality.  See
Zhao and Kit~\cite{zhao2008empirical} for a detailed review.

Nonparametric Bayesian methods have been introduced to unsupervised word
segmentation by Goldwater et al.~\cite{goldwater2009bayesian} to address the problem of
context modeling.  They relied on a hierarchical Dirichlet process (HDP) to
better capture between-word dependencies and avoid misidentifying common word
collocations as whole words.  This approach was later extended by
Mochihashi et al.~\cite{mochihashi2009bayesian} that installs two Pitman-Yor processes at both
word and character levels, aiming at a joint resolution of the dependency
issues.  Johnson and Goldwater~\cite{johnson2009improving} proposed an extension of HDP on
probabilistic context-free grammar, called adaptor grammars, whose performance
is currently state of the art for unsupervised phonemic word segmentation.

Minimum description length (MDL) is an inductive principle in information
theory that equates salience to parsimony in statistical modeling
\cite{rissanen1978modeling,rissanen1983universal}.  It is perhaps best known
for the connection between the famous two-part code and the Bayesian formalism.
Early applications of MDL in computational linguistic target mainly on very
general unsupervised learning problems, such as grammar induction
\cite{grunwald1996minimum} and language acquisition
\cite{deMarcken1996linguistic,deMarcken1996unsupervised,brent1996distributional,brent1999efficient}.
There has been some constant interest from problem subareas in applying MDL to
their work, but in the early days MDL served mainly as a motive for Bayesian
modeling.  Not much work has been done to explore the true merit of MDL as an
\emph{information criterion} until recently in lexical acquisition subareas
such as morphology \cite{goldsmith2001unsupervised,creutz2002unsupervised} and
unsupervised word segmentation \cite{yu2000unsupervised,argamon2004efficient}.

%--------------------------------------------------
% In the early development of unsupervised word segmentation, MDL has played a
% crucial role for motivating probabilistic modeling.  The notion of two-part MDL
% code introduced by \cite{rissanen1978modeling} has quickly become an equivalent
% formalism to Bayesian learning.  It has invited further algorithmic work on
% distributional assumptions that underlie language.  
%-------------------------------------------------- 

%--------------------------------------------------
% In grammar induction,
% \newcite{grunwald1996minimum} proposed a simple algorithm for learning grammar
% rules that best compress the data.  Similar approaches were taken by other work
% in lexical acquisition and morphology.
% \newcite{brent1996distributional,brent1999efficient} presented an analysis
% based on distributional regularities to word boundary discovery, using an
% exhaustive search over the entire hypothesis space.  De Marcken
% \shortcite{deMarcken1996linguistic,deMarcken1996unsupervised} used an EM
% algorithm to approximate the maximum a posteriori learning for the lexicon.
% \newcite{goldsmith2001unsupervised} developed an MDL-based morphology analyzer
% called \emph{linguistica}.  \newcite{creutz2002unsupervised} applied a more
% effective MDL formulation to their famous Morfessor system.
%-------------------------------------------------- 

% \newcite{yu2000unsupervised} is among the first to apply model selection to
% unsupervised word segmentation.  

\begin{figure*}[t]
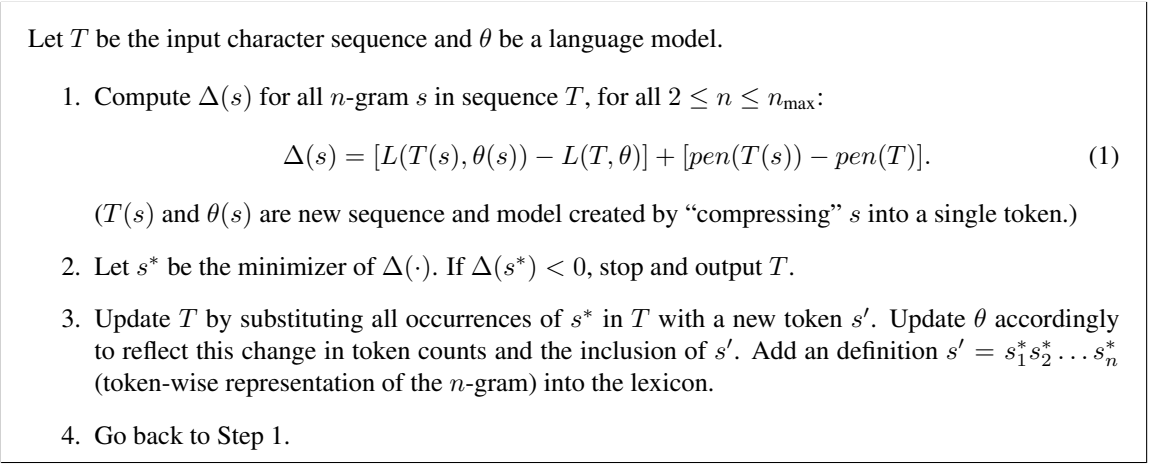

  \begin{mdframed}
    
    Let $T$ be the input character sequence and $\theta$ be a language model.

    \begin{enumerate} 
      \item Compute $\Delta(s)$ for all $n$-gram $s$ in sequence $T$, for all $2 \le n \le n_\text{max}$:
      \begin{equation} \Delta(s) = [L(T(s), \theta(s)) - L(T, \theta)] + [pen(T(s)) - pen(T)]. \end{equation}
	($T(s)$ and $\theta(s)$ are new sequence and model created by
	``compressing'' $s$ into a single token.)

      \item Let $s^*$ be the minimizer of $\Delta(\cdot)$.  If $\Delta(s^*) < 0$, stop and output $T$.  
      \item Update $T$ by substituting all occurrences of $s^*$ in $T$ with a new
	token $s'$.  Update $\theta$ accordingly to reflect this change in token
	counts and the inclusion of $s'$.  Add an definition $s' = s^*_1 s^*_2 \ldots s^*_n$ (token-wise
	representation of the $n$-gram) into the lexicon.  
      \item Go back to Step~1.
    \end{enumerate}
  \end{mdframed}

  \caption{The incremental learning framework.}
  \label{f:algorithm}
\end{figure*}

This idea has later been extended in
Zhikov et al.~\cite{zhikov2010efficient}.  They varied the decision threshold on branching
entropy to generate a number of candidate segmentations, each would later go
through an adjustment procedure for locally reducing the description length.
All these candidates are then judged using the MDL criterion to produce the
final output.  This method works well on both phonemic and orthographic
corpora, and has gather some attention from the research community.  A number
of follow-up work has devoted to experimenting with different boundary placing
strategies.  Hewlett and Cohen~\cite{hewlett2011fully} used Bootstrapped Voting Experts, a
sophisticated algorithm also based on branching entropy, to take the job of
segmentation generation.  Chen et al.~\cite{chen2012regularized} proposed using
regularized compression, a compression-based algorithm that builds up word
representation from characters in a bottom-up fashion; a later version of this
algorithm in Chen~\cite{chen2013improved} had been shown to work well on phonemic
data.  Recently, some negative result with MDL on orthographic Chinese
segmentation was reported in Magistry and Sagot~\cite{magistry2013mdl}, suggesting that MDL has
sometimes failed to improve their baseline measure.

\newcommand{\AIC}[1]{\ensuremath{\text{AIC}_{#1}}}
\newcommand{\MDL}[1]{\ensuremath{\text{MDL}_{#1}}}

\section{Incremental Learning Using Maximum Penalized Likelihood}\label{s:incremental-learning}

There has been some efforts aiming at formalizing the MDL-based approach for
unsupervised word segmentation \cite{zhikov2010efficient,chen2012regularized}.
In this section, we propose a general framework that incorporates many design
elements from these previous attempts.  We chose between two types of search
strategies: in a space of boundary placements \cite{zhikov2010efficient} or a
space spanned by lexemes in \cite{chen2012regularized}.  As our goal is to have
something simple to work on, eventually we settle on Chen et al.'s method for
ease of implementation.

\paragraph{Framework}  Our algorithm is given in Figure~\ref{f:algorithm}.  Let
us first assume the input to algorithm is a sequence of $N$ characters, denoted
as $X^N$.  The basic idea of incremental learning is to iteratively process
this sequence of ``words'' and make it more compact.  Initially, this sequence
contains only characters, $t_1 t_2 \ldots t_N = X^N$.  Then, in each iteration
our algorithm would pick a contiguous subsequence of words from the input, say
$s = t_i t_{i+1} \ldots t_{i+n-1}$ (for some $i$ and $n \le n_\text{max}$), and
concatenate all its components into one single string.  We do this for
\emph{all} occurrences of $s$, thereby reducing the total number of words in
the original word sequence.

For simplicity let us write the input sequence as $T$ and the new sequence
produced by compressing some candidate $s$ as $T(s)$.  In each iteration, we
look for some candidate $s^*$ that minimizes \begin{equation} L(T(s),\theta(s))
+ pen(T(s)), \label{e:objective} \end{equation} where $L(T, \theta)$ is the sum
of the log likelihood estimate and the complexity term for $\theta$, as in:
\begin{equation} - \log p(T|\theta) - \frac{1}{2} (\text{\# unigram}) \log N.
\end{equation} Note that $p(T|\theta)$ is a categorical distribution over
unigram words; we use maximum likelihood to estimate this probability.

Since our goal is to find the candidate that minimizes the change plus the
penalty, it suffices in each iteration to calculate only the change in the
penalized maximum likelihood estimate.  We have implemented efficient update
steps similar to Chen~\cite{chen2013improved} by deriving a bound for the
likelihood component using the mean-value theorem.  These steps are omitted
here for space limitation.

\paragraph{Penalized Likelihood}  

% The objective function in Chen et al. has
% caused some difficulty in analysis since it depends on parameters that are
% notoriously hard to get right, such as compression rate (to decide when to
% stop) and minimum support (to filter out low-frequency signals).

Our objective function \eqref{e:objective} in the incremental learning
algorithm can be seen as a joint log-likelihood estimate $p(X^N, T, \theta)$.
The first component $p(T, \theta)$ has been accounted for in $L(T, \theta)$;
what was left is $p(X^N | T, \theta)$, the probability of generating the input
based on a sequence of words $T$ and a language model $\theta$.  This is not
about testing whether $T$ is a feasible segmentation for $X^N$, but more about
assessing how likely one may think the segmentation is reasonable.  Inspired by
the string-length-based priors used in Bayesian modeling
\cite{goldwater2009bayesian}, we assign this component as a penalty prior for
demoting long word formation.  This penalty term is by design a summation of
\emph{super-additive} function values of individual token length (in
characters).  A super-additive function $f$ satisfies that $f(x + y) > f(x) +
f(y)$ for all $x$ and $y$.  This definition aims to approximate the cognitive
overhead created by composition: putting more sub-word units together would
create new meanings.  This is best illustrated by considering the number of
possible collocations and singletons in $n$ word units, which is a quadratic
function $\binom{n}{2} + \binom{n}{1}$, or more naively the number of possible
combinations $2^n$.  Both examples here are in super-additive forms.

In this paper, we consider the following two types of penalties, $x \log x$ and
$x^2$.  We also find it useful to have an intercept term.  The penalties are
defined as follows: \begin{equation} \begin{split} pen_1(T) &= - \alpha |t| +
\beta \sum_{t \in T} |t| \log |t|, \\ pen_2(T) &= - \alpha |t| + \beta \sum_{t
\in T} |t|^2. \end{split} \end{equation}

\paragraph{Model Selection}  In this paper, we use Akaike information criterion
(AIC) and minimum description length (MDL) to measure the quality of a specific
segmentation output.  These criteria are applied to data-model
pair $(T, \theta)$.  For AIC, we use the finite correction proposed by
Sugiura~\cite{sugiura1978further}:
\[ \text{AIC}_\text{c} = - \log p_{\hat\theta}(X^N) + 
\frac{N k}{N - k - 1},\]
where $p_{\hat\theta}(X^N)$ is the maximum
likelihood estimate for observations $X^N$ and $k$ is the complexity term (number of parameters
needed for fitting the model $\hat\theta$.)  
For the definition of MDL, we follow Rissanen~\cite{rissanen1983universal} and add a
complexity term for coding the word-level codebook, as in
Zhikov et al.~\cite{zhikov2010efficient}.  Our final formula is defined as: \[ \text{MDL}
= - \log p_{\hat\theta}(X^N) + \frac{k}{2} \log N + cbl(\theta), \]  
where $cbl(\theta)$ is the code length of the lexicon.  Both criteria depends
on a log-likelihood component, whose estimation is based on the designated
language model $\theta$.  In this study, we consider unigram, bigram, and
trigram models.

% \newcite{akaike1974new} introduced a famous equation, later called Akaike
% information criterion (AIC), intended as
% asymptotic estimate of the Kullback-Leibler divergence from the estimated model
% to the ``true'' distribution.  It has a problem that model complexity will be underestimated when the observation is of finite
% length, therefore 

%--------------------------------------------------
% \newcite{schwarz1978estimating} proposed another formula motivated in the
% Bayesian context, known as Bayesian information criterion (BIC).
% Interestingly, at about the same time \newcite{rissanen1978modeling} derived
% the first instantiation of his famous two-part code, which was later shown by
% \newcite{rissanen1983universal} to be equivalent to the BIC formula.  This
% criterion, which we called MDL just to be consistent with the literature, is
% defined as: \[ \text{MDL} = -2 \log p_{\hat\theta}(X^N) + k \log N. \]  
%-------------------------------------------------- 

\section{Experiment}\label{s:experiment}

Our incremental learning algorithm relies on two free parameters $\alpha$ and
$\beta$ to assign correct penalty weights to each segmentation candidate.  The
best combination may depend on language, representation, and even corpus
statistics.  It is therefore infeasible to assume one magic setting that works
well universally.  In this experiment, we show how the adaption to data can be
achieved using model selection.  Furthermore, we justify how super-additive
penalty may improve incremental learning, and also verify the influence of
higher-order dependencies to model selection.  Standard performance measures
such as precision (P), recall (R) and F-measures (F) are used for evaluating
segmentation accuracy.  We report these figures at three levels: token,
boundary, and lexicon.  The experimental setup and our methodology are detailed
as follows.

% latex table generated in R 3.0.2 by xtable 1.7-1 package
% Thu Mar 20 01:17:14 2014
\begin{table*}[t]
\centering
\begin{tabular}{lc|cc|ccccccccc}
  & & $\alpha$ & $\beta$ & P & R & F & BP & BR & BF & LP & LR & LF \\ 
  \hline
  \multicolumn{2}{l|}{Base} & 0.0 & 0.0 & 37.9 & 14.5 & 21.0 & 95.6 & 12.2 & 21.7 & 9.8 & 40.1 & 15.8  \\
  \multicolumn{2}{l|}{Base (stop in 500 runs)} & 0.0 & 0.0 & 51.0 & 54.3 & 52.6 & 68.7 & 75.1 & 71.8 & 48.8 & 18.8 & 27.1 \\
  \hline \hline
  % & & $\alpha$ & $\beta$ & P & R & F & BP & BR & BF & LP & LR & LF \\ 
  % \hline
  $\text{AIC}_1$ & 229474.0     & 1.6 & 0.4 & 38.5 & 17.2 & 23.8     & 99.0 & 21.5 & 35.4     & 11.8 & 40.8 & 18.3 \\ 
  $\text{AIC}_2$ & 178471.4     & 4.9 & 1.3 & 46.1 & 27.0 & 34.0     & 98.9 & 40.9 & 57.9     & 25.9 & 63.8 & 36.8 \\ 
  $\text{AIC}_3$ & \bf 150167.8 & 1.9 & 1.4 & 82.6 & 78.6 & \bf 80.5 & 92.9 & 86.5 & \bf 89.6 & 62.4 & 61.6 & \bf 62.0 \\ 
  \hline
  $\text{MDL}_1$ & 323797.8     & 0.5 & 0.7 & 71.8 & 57.9 & 64.1     & 94.7 & 68.8 & 79.7     & 49.8 & 64.3 & 56.2 \\ 
  $\text{MDL}_2$ & \bf 290370.0 & 3.7 & 2.3 & 83.2 & 84.2 & \bf 83.7 & 90.5 & 92.1 & \bf 91.3 & 59.2 & 56.3 & \bf 57.7 \\ 
  $\text{MDL}_3$ & 320373.9     & 1.5 & 1.5 & 73.0 & 79.8 & 76.2     & 82.7 & 93.5 & 87.8     & 59.5 & 46.7 & 52.3 \\
  \hline \hline
  % & & $\alpha$ & $\beta$ & P & R & F & BP & BR & BF & LP & LR & LF \\ 
  % \hline
  $\text{AIC}_1$ & 237542.5     & 3.7 & 0.1 & 27.8 & 12.4 & 17.2     & 99.2 & 21.5 & 35.4     &  8.9 & 35.8 & 14.3 \\ 
  $\text{AIC}_2$ & 177499.9     & 3.7 & 0.2 & 45.2 & 27.0 & 33.8     & 98.5 & 42.5 & 59.3     & 26.0 & 65.6 & 37.2 \\ 
  $\text{AIC}_3$ & \bf 150730.3 & 0.6 & 0.3 & 81.6 & 78.5 & \bf 80.0 & 91.5 & 86.6 & \bf 89.0 & 61.2 & 61.2 & \bf 61.2 \\ 
  \hline
  $\text{MDL}_1$ & 326346.4     & 0.0 & 0.2 & 81.3 & 74.4 & 77.7     & 94.3 & 82.9 & 88.3     & 61.6 & 63.1 & \bf 62.4 \\ 
  $\text{MDL}_2$ & \bf 290033.9 & 1.7 & 0.5 & 79.7 & 81.1 & \bf 80.4 & 88.0 & 90.1 & \bf 89.0 & 58.3 & 55.1 & 56.6 \\ 
  $\text{MDL}_3$ & 319573.5     & 0.0 & 0.3 & 72.5 & 80.1 & 76.1     & 82.1 & 94.3 & 87.8     & 60.4 & 46.6 & 52.6 \\ 
\end{tabular}
\caption{Test results on the Bernstein-Ratner corpus for different incremental
learning settings: zero penalty (top), $x \log x$ penalty (middle) and $x^2$
penalty (bottom). } \label{t:performance}
\end{table*}

\subsection{Phonemic Word Segmentation}\label{s:phonemic-word-segmentation}  

We used the Brent's version of Bernstein-Ratner corpus to evaluate our method.
\cite{brent1996distributional,bernstein-ratner1987phonology}.  This corpus is
the phonemic transcription of English child-directed speech from the CHILDES
database \cite{macwhinney1990child}, and has been widely used as a standard
testbed for unsupervised word segmentation.  It has 9,790 utterances that
comprise totally 95,809 words in phonemic representation.  

Our result will be compared with the following methods: (1) incremental
learning without adding the penalty, i.e., setting $\alpha = 0.0$ and $\beta =
0.0$, (2) MDL-based methods such as regularized compression
\cite{chen2013improved} and EntropyMDL \cite{zhikov2010efficient}, and (3)
adaptor grammars \cite{johnson2009improving}.  

We used two classes of information criteria
in this experiment, derived from AIC and MDL.  Both classes depend on a
likelihood term $- \log P(X^n)$, for which in this study we obtained three
estimates under the unigram, bigram, and trigram assumptions.  We denote these
criteria as \AIC{n} and \MDL{n} respectively for $n = 1, 2, 3$.  The complexity
term $K$ (degrees of freedom) for \AIC{n} is given by:
\begin{equation}
  \begin{split}
    \AIC{1} &= \sum_w (1 + |w|) + \text{\# unigrams}, \\
    \AIC{2} &= \sum_w (1 + |w|) + 1 + 2 \times \text{\# bigrams}, \\
    \AIC{3} &= \sum_w (1 + |w|) + 1 + 2 \times \text{\# trigrams}.
  \end{split}
\end{equation}
For \MDL{n}, we have the following formula:
\begin{equation}
  \MDL{n} = \, \text{\# $n$-grams}
\end{equation}

We ran a grid search for both $\alpha$ and $\beta$ from 0 to 5 with step size
0.1, which amounts to 2,601 combinations in total.  For each combination, we
first let the algorithm run to finish, collect the output, and then compute the
AIC and MDL values.  This procedure is repeated on both penalty settings.  Each
AIC or MDL criteria will have one combination that achieves the minimum.  We
collected these combinations and then chose the one that minimize over the
\emph{entire family} as the designated output for \AIC{n} or \MDL{n}.

\paragraph{Result} The test result is summarized in Table~\ref{t:performance}.
From top to bottom we have the results using different penalties: zero
penalty, $x \log x$, and $x^2$.  Zero-penalty and its upper bound (the best
possible result among intermediate output) are included here merely as lab
controls for the super-additive penalties.  Zero penalty
itself did not do very well, achieving only 21.6 for token F-score (hereafter
abbreviated as F-score or F-measure); its best intermediate output was found in
500 iterations, giving a mediocre 52.6 {in} F-measure.

With $x \log x$ penalty, \AIC{n} has achieved 23.8, 34.0, and 80.5 {in}
F-score, respectively for $n = 1, 2, 3$.  \MDL{n} does slightly better in
general, giving 64.1, 83.7, and 76.2 {in} F-score.  The best performance for
AIC and MDL is on \AIC{3} and \MDL{2}.  These two runs are still clear winners
on $x^2$ penalty.  In F-measure, \AIC{3} and \MDL{2} does equally well,
delivered 80.0 and 80.4 respectively {in} terms of F-measure.  The performance
for both criteria on $x^2$ is very close to \AIC{3} on $x \log x$.  

\begin{table*}[t]
\centering
\begin{tabular}{lccccccccc}
  & P & R & F & BP & BR & BF & LP & LR & LF \\ 
  \hline
  $\text{AIC}_3$ & 82.9 & 79.4 & 81.1     & 92.7 & 87.2 & 89.9     & 62.7 & 63.1 & \bf 62.9 \\ 
  $\text{MDL}_2$ & 83.7 & 84.2 & \bf 84.0 & 91.0 & 91.8 & \bf 91.4 & 59.9 & 58.0 & 58.9 \\ 
  \hline
  $\text{AIC}_3$ & 82.5 & 81.0 & \bf 81.7 & 91.2 & 88.9 & \bf 90.0 & 59.9 & 61.2 & \bf 60.5 \\ 
  $\text{MDL}_2$ & 80.7 & 80.6 & 80.6     & 89.5 & 89.3 & 89.4     & 59.3 & 57.8 & 58.6 \\ 
\end{tabular}

\caption{Performance results for top-10 ensemble using $x \log x$ (top) and $x^2$
(bottom) penalties. } \label{t:ensemble}
\end{table*}

\begin{table*}[t]
  \begin{minipage}{0.49\textwidth}
    \centering
    \begin{tabular}{lcccc}
      & P & R & F \\
      \hline
      $\text{MDL}_2$, $x \log x$ + ensemble             & 83.7 & 84.2 & \bf 84.0 \\
      $\text{AIC}_3$, $x^2$ + ensemble                  & 82.5 & 81.0 & 81.7 \\
      Chen \cite{chen2013improved}                      & 79.1 & 81.7 & 80.4 \\
      Hewlett and Cohen \cite{hewlett2011fully}         & 79.3 & 73.4 & 76.2 \\
      Zhikov et al. \cite{zhikov2010efficient}          & 76.3 & 74.5 & 75.4 \\
    \end{tabular}
  \end{minipage}
  \begin{minipage}{0.49\textwidth}
    \centering
    \begin{tabular}{lcccc}
      & F \\
      \hline
      Adaptor grammars, colloc3-syllable      & \bf 87.0 \\
      $\text{MDL}_2$, $x \log x$ + ensemble   & 84.0 \\
      $\text{AIC}_3$, $x^2$ + ensemble        & 81.7 \\
      Adaptor grammars, colloc                & 76.0 \\
      Adaptor grammars, unigram               & 56.0 \\
    \end{tabular}
  \end{minipage}

  \caption{Performance comparison with MDL-based methods (left) and adaptor
  grammars (right).  Figures for the latter were reproduced from the reference
  implementation \cite{johnson2009improving} using batch initialization and
  maximum marginal decoding.  Note that colloc3-syllable adaptor grammars is
  not fully unsupervised due to a small amount of phoneme productions built
  into its core.} \label{t:comparison}

\end{table*}

In general, the trigram estimate \AIC{3} works the best for AIC, constantly
achieving the least decision value among the three and the best performance.
The same goes for the bigram estimate \MDL{2} in the MDL camp, whose
performance surpassed the other three consistently across two penalty settings.
Furthermore, the unigram and bigram estimates do not seem to play well with
AIC.  On either penalty setting, both \AIC{1} and \AIC{2} have struggled to
catch up with the zero-penalty upper-bound performance.  The result on MDL runs
seem more coherent in performance.  \MDL{1} and \MDL{3} lagged behind \MDL{2}
only by a small margin about 3 to 4 points in F-score.  From the test result,
it seems fair to conclude that the proposed incremental learning framework is
effective in unsupervised word segmentation.

We also experimented with a simple ensemble method that combines top-$k$
segmentation outputs.  Suppose there are totally $n$ plausible positions for
placing boundaries on the input corpus.  For each position, we check how many
in the $k$ outputs have actually placed a boundary there and seek to obtain a
majority decision.  These $n$ decisions are later put together as a combined
output; ties are interpreted as ``no-boundary-here''.  We tested top-10
ensemble for \AIC{3} and \MDL{2} on both penalty settings.  The result is given
in Table~\ref{t:ensemble}.  The token-level performance for all four
experimental runs was improved by 0.2 to 1.7 points {in} F-score.  Top-10
ensemble also slightly improves lexicon F-measure on $x \log x$ penalties,
although in general it has a mixed effect on the other measures.  The overall
best performance is now on \MDL{2} using $x \log x$ penalty, achieving 84.0
{in} token F-score.

In Table~\ref{t:comparison}, we make an overall comparison between our result
and reference methods, including MDL-based methods and adaptor grammars.  The
summary on the left shows that our incremental learning framework has
outperformed all the existing MDL-based approaches, winning out regularized
compression by about 4 points in F-score.  On the right, we find the
performance of incremental learning has surpassed both unigram and colloc
adaptor grammars by a large margin.  We also compared our approach with
colloc3-syllable adaptor grammars, which is commonly thought as weakly
supervised.  The result shows that incremental learning approach lagged behind
colloc3-syllable by 3 points in F-score.  Our interpretation is that this more
advanced version of adaptor grammars has built in some linguistic/structural
assumptions that have no equivalent in our simplistic framework.  In the case of
colloc3-syllable, it was a small set of phonemic productions that helps
improving syllable-level modeling accuracy.  

Despite the inability of modeling sophisticated generative nature in language,
incremental learning has good processing speed that permits more CPU cycles to
go into parameter estimation.  On our test machine, it took merely 3 to 4
seconds for testing one combination of $\alpha$ and $\beta$.  Running 7-fold
colloc3-syllable adaptor grammars would take roughly 4 days; our approach would
finish in 3 hours even on a $51 \times 51$ grid search.  Detailed results are
given in Table~\ref{t:running-time}. 

\begin{table}[t]
  \centering
  \begin{tabular}{lr}
    & Time (s)\\
    \hline
    $\text{MDL}_2$, $x \log x$ + ensemble, 51 $\times$ 51 & \bf 9,350 \\
    Adaptor grammars, colloc & 73,356 \\
    Adaptor grammars, colloc3-syllable &  376,732 \\
  \end{tabular}

  \caption{Timing results on the Bernstein-Ratner corpus, all methods tested on
  an Intel Xeon 2.5GHz 8-core machine with 8GB RAM.  All the tests were done on
  a single core; no parallelization was intended in this timing benchmark.}

  \label{t:running-time}
\end{table}

\begin{table}[t]
\centering
\begin{tabular}{lcccc}
  & AS & MSR & CityU & PKU \\
  \hline
  $\text{MDL}_2$, $x \log x$                & \bf 80.9 & \bf 80.0 & 80.5     & \bf 80.6 \\
  Wang et al. \cite{wang2011new}, Setting 3 & 76.9     & 79.7     & \bf 80.8 & 79.3 \\
  Zhikov et al. \cite{zhikov2010efficient}  & --       & 79.5     & 79.8     & -- \\
  Chen et al. \cite{chen2012regularized}    & --       & 77.4     & 77.0     & -- \\
  Zhao and Kit \cite{zhao2008empirical}     & --       & 66.5     & 68.4     & -- \\
\end{tabular}

\caption{Test results in token F-measure on the SIGHAN Bakeoff-2005 training
sets.  The baseline results all come from the literature.}
\label{t:comparison2}

\end{table}

\subsection{Orthographic Word Segmentation}\label{s:orthographic-word-segmentation}

We also tested our approach under the setting for orthographic word
segmentation on a public corpus SIGHAN Bakeoff-2005
\cite{emerson2005second}.  This corpus is a standard benchmark for Chinese word
segmentation; it has 4 subsets, and each comes with a official training/test
split.  For simplicity, our experiment was conducted exclusively on these
training sets.  The details are given below.  

\begin{center}
  \begin{tabular}{lrr}
    & \# passages & \# words \\
    \hline
    AS & 708,953 & 5.45M \\
    MSR & 86,924 & 2.37M \\
    CityU & 53,019 & 1.46M \\
    PKU & 19,056 & 1.10M 
  \end{tabular}
\end{center}

It is easy to see that each of these sets is much more sizable than our
previous experiment.  As of this writing, it remains difficult if not
impossible to apply hierarchical Bayesian methods to a corpus of this size.
Hence, in this experiment, we chose to compare with less sophisticated
approaches such as MDL-based methods, DLG \cite{zhao2008empirical}, and ESA
\cite{wang2011new}.  The latter is known to have the best segmentation accuracy
on the Bakeoff-2005 corpus, although the best figures reported was achieved by
explicitly hand-picking the parameter.  To make this a fair comparison, we
instead took the median from all the reported runs for 10-iteration ESA.  Note
that ESA still has the best performance after this fix.  Our experiment setup
is compatible with the setting 3 for ESA, meaning that hard boundaries are set
up around punctuation marks.  Besides that, little preprocessing was done to
the corpus body.

Due to the scale of experiment, we chose to test only the best setting found in
the previous round: $\text{MDL}_2$ with $x \log x$ penalty.  Even so, the
experiment remains very time-consuming; on the largest set AS, it took several
hours for our incremental learning algorithm to complete.  Therefore, for
estimating the parameters we first ran a search on $\alpha$ and then a second
on $\beta$ in a way analogous to the previous experiment.  We ran this grid
search only once on the AS set; the found combination $(\alpha, \beta) = (2.0,
3.0)$ was used throughout the rest of the experiment on other subsets.  

\begin{figure*}[t]
  \centering 
  \begin{minipage}{0.275\textwidth}
    \includegraphics[width=\linewidth]{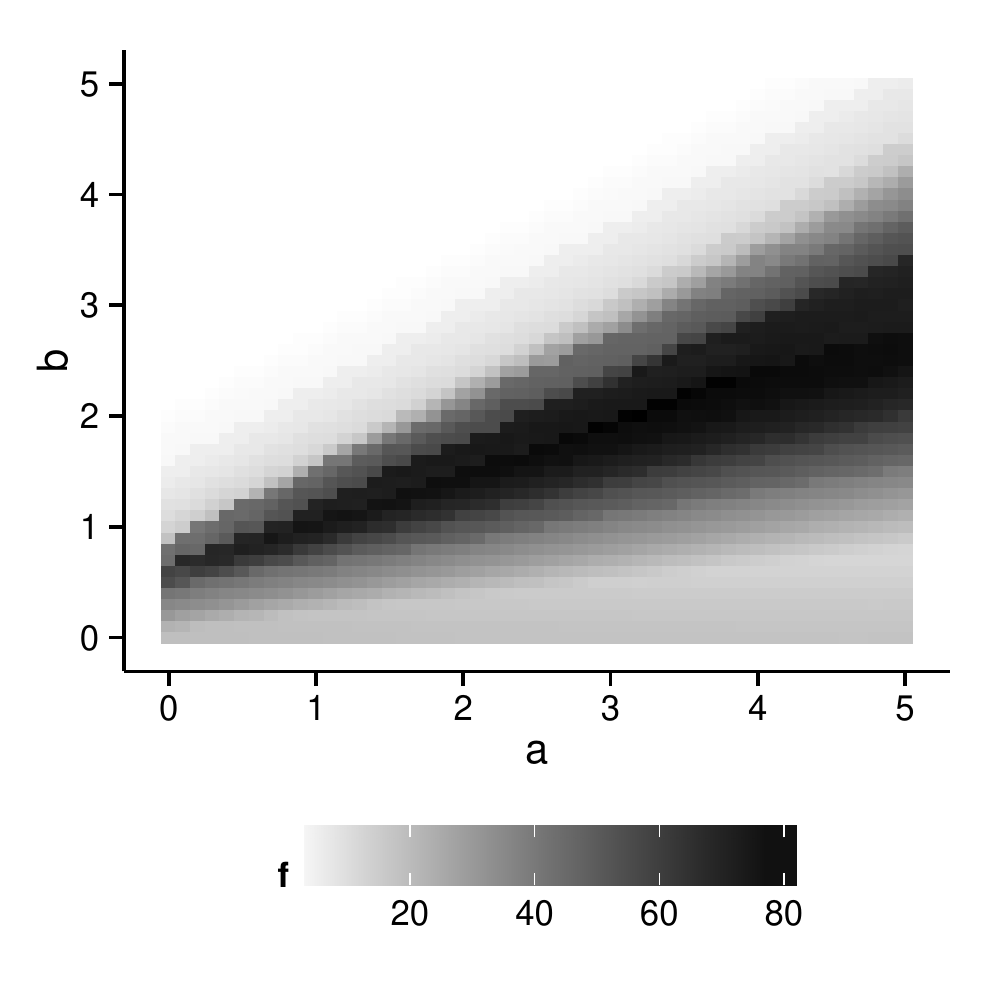}
  \end{minipage}
  \begin{minipage}{0.275\textwidth}
    \includegraphics[width=\linewidth]{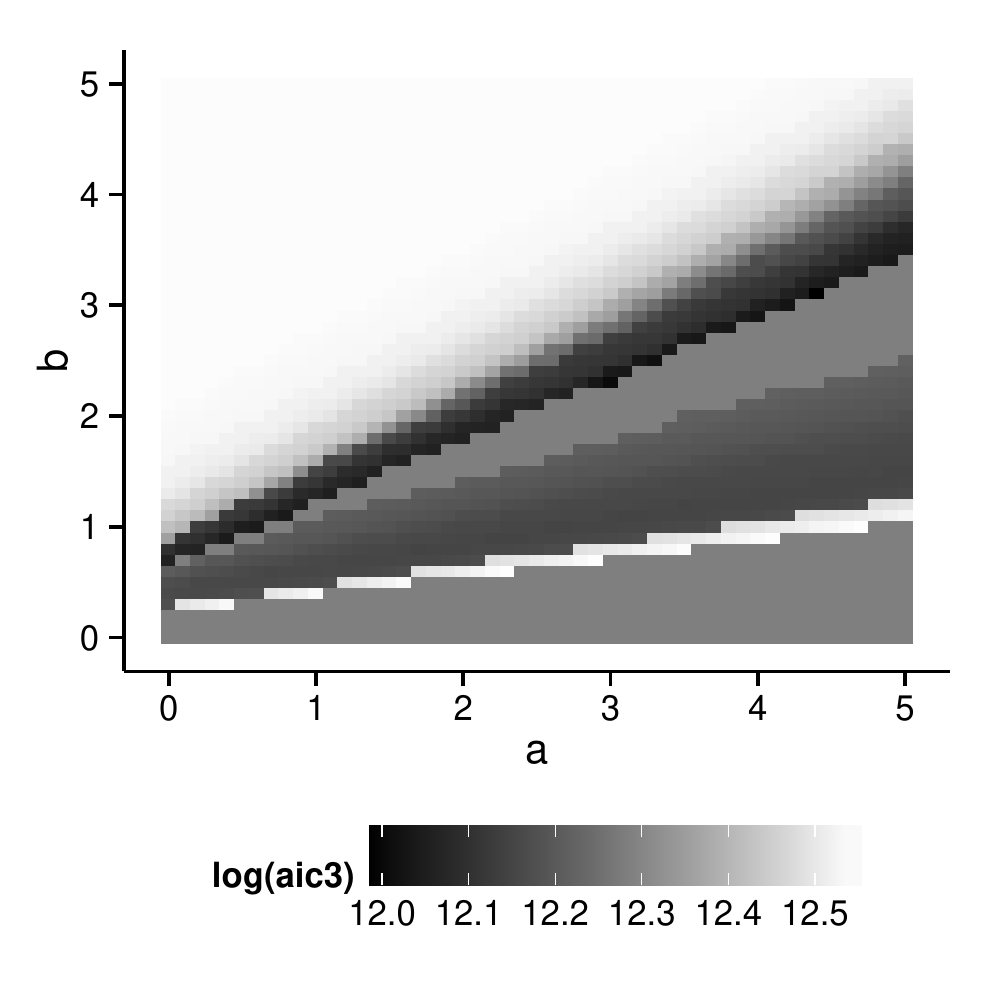}
  \end{minipage}
  \begin{minipage}{0.275\textwidth}
    \includegraphics[width=\linewidth]{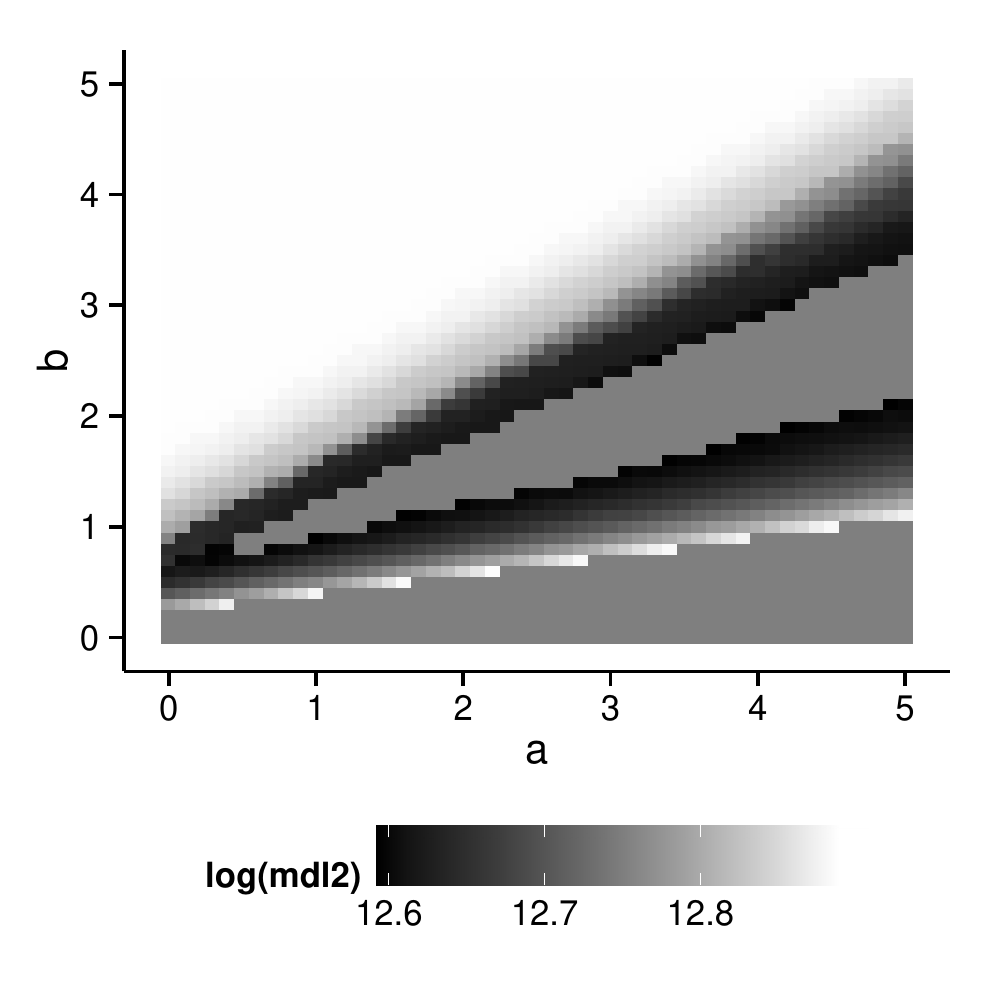}
  \end{minipage}

  \begin{minipage}{0.275\textwidth}
    \includegraphics[width=\linewidth]{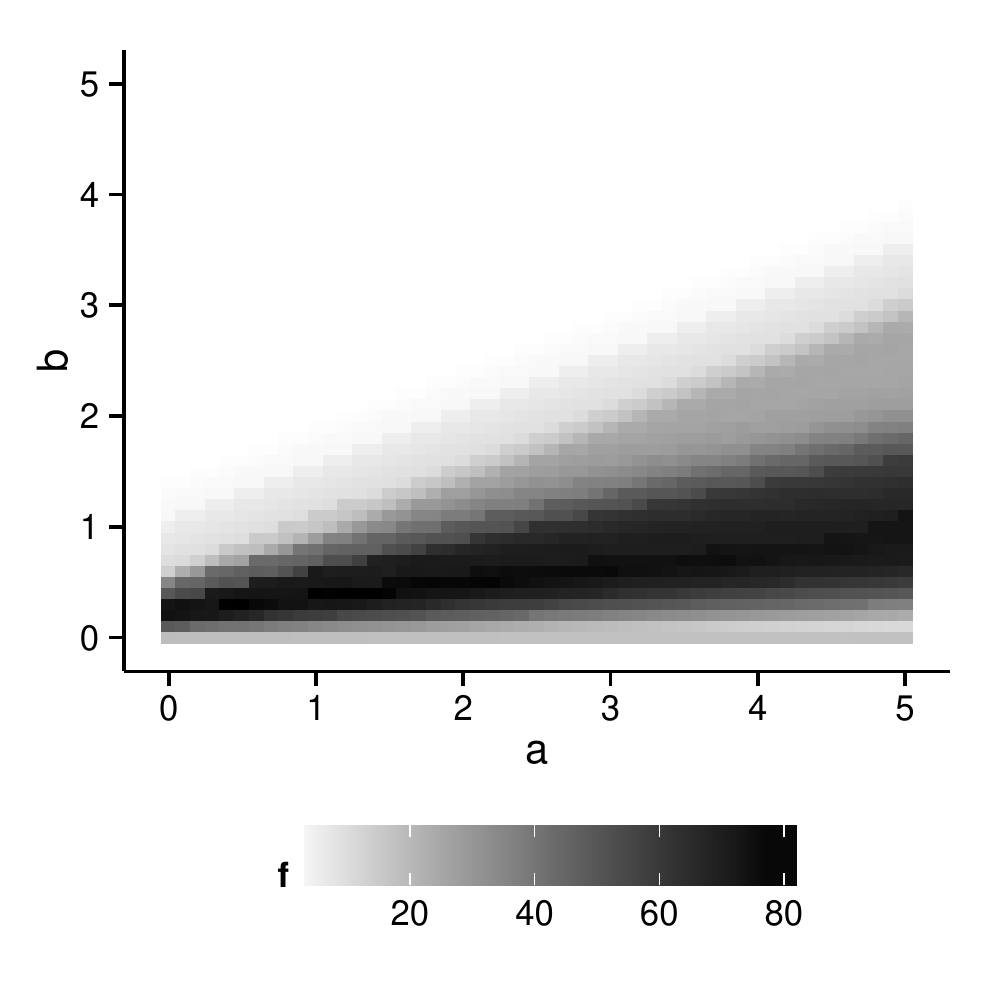}
  \end{minipage}
  \begin{minipage}{0.275\textwidth}
    \includegraphics[width=\linewidth]{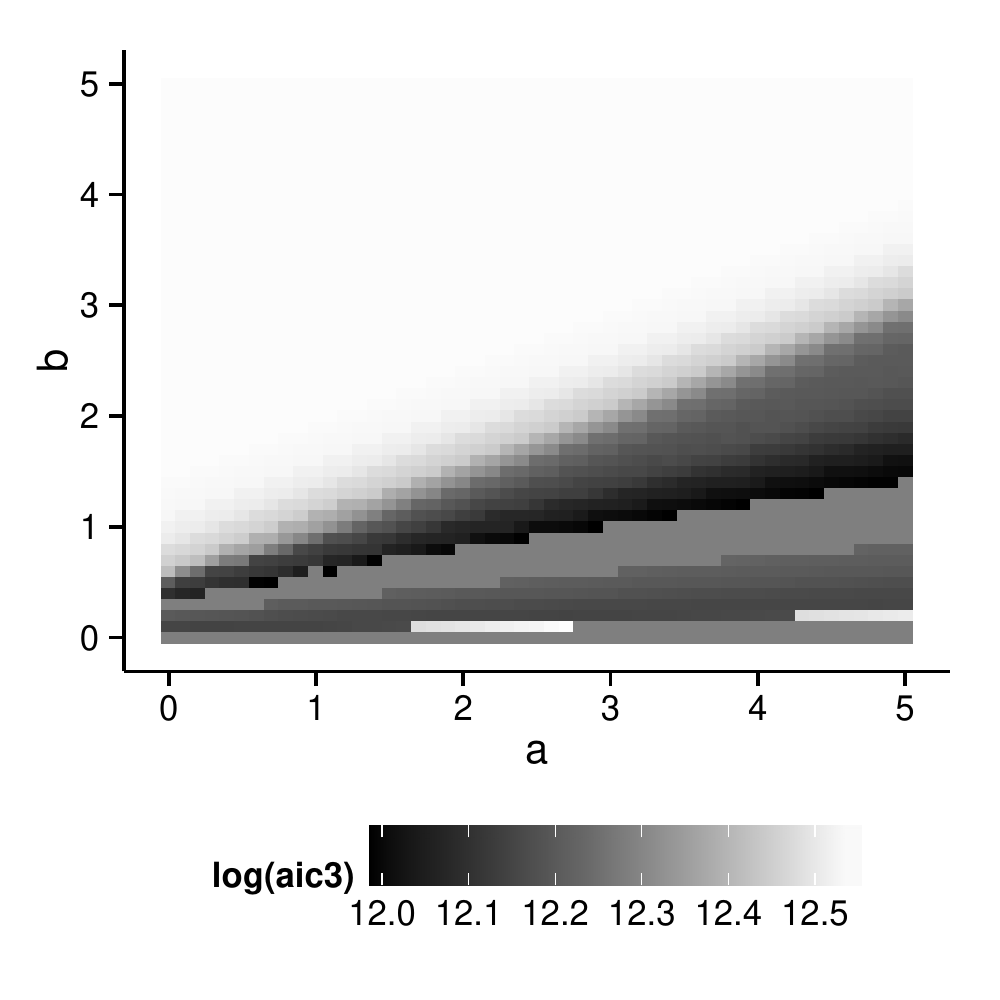}
  \end{minipage}
  \begin{minipage}{0.275\textwidth}
    \includegraphics[width=\linewidth]{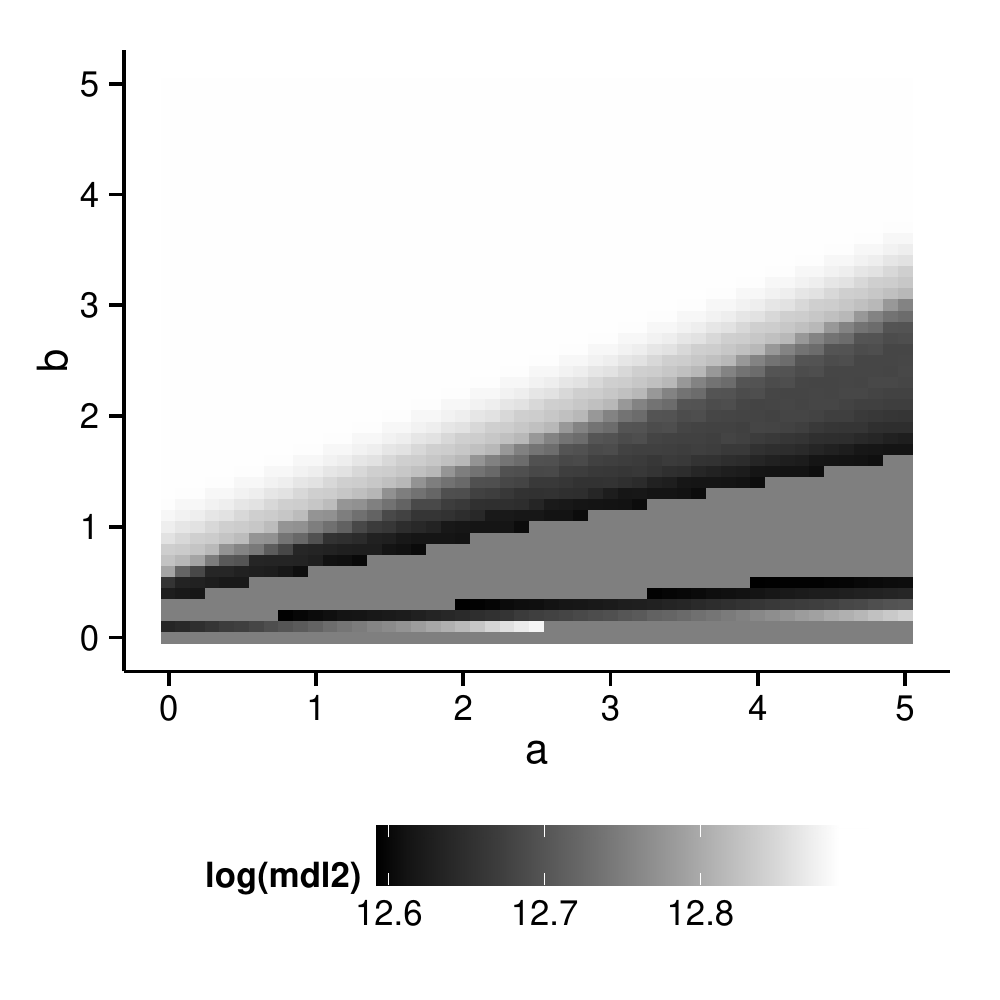}
  \end{minipage}

  \caption{Heat maps printed in greyscale for token F-score (left), \AIC{3}
  (middle), and \MDL{2} (right) in a truncated hypothesis space represented by
  $\alpha$ and $\beta$, based on $x \log x$ (top) and $x^2$ (bottom) penalties.
  Note that \AIC{3} and \MDL{2} are log-transformed for better contrast.}
  \label{f:heatmap}

\end{figure*}

% latex table generated in R 3.0.2 by xtable 1.7-1 package
% Thu Mar 20 00:26:52 2014
\begin{table*}[t]
\centering
\begin{tabular}{lccccccc}
  & Penalty & $\text{AIC}_1$ & $\text{AIC}_2$ & $\text{AIC}_3$ & $\text{MDL}_1$ & $\text{MDL}_2$ & $\text{MDL}_3$\\ 
  \hline
  \multirow{2}{*}{Output} & $x \log x$  & -0.80 & -0.79 & -0.76 & -0.93 & -0.78 & -0.46 \\ 
  & $x^2$  & -0.94 & -0.89 & -0.87 & -0.90 & -0.87 & -0.54 \\ 
  \hline
  \multirow{2}{*}{Full trace} & $x \log x$ & 0.44 & -0.47 & -0.79 & -0.79 & -0.96 & -0.92 \\ 
  & $x^2$ & 0.03 & -0.53 & -0.78 & -0.73 & -0.93 & -0.86 \\ 
\end{tabular}
\caption{Rank correlations (Spearman's $\rho$) between token F-measure and information criteria}
\label{t:correlation}
\end{table*}

\begin{figure}[t]
  % \fakebox[0.9 \columnwidth]{3in}
  \centering
  \includegraphics[width=0.5\linewidth]{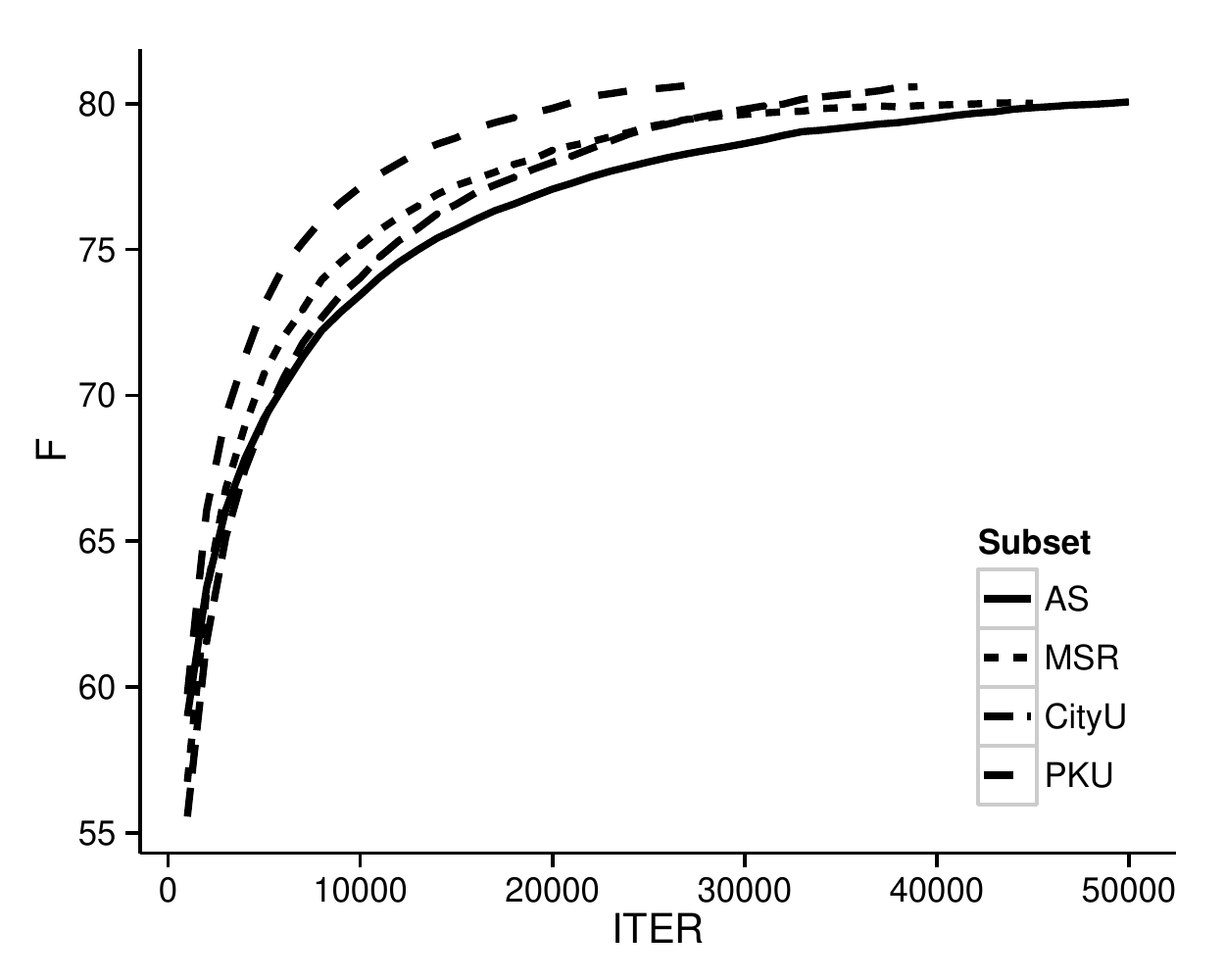}

  \caption{The improvement on F-score across iterations on the Bakeoff-2005 corpus.}
  \label{f:improvement}
\end{figure}

\paragraph{Result}  We had our method run to stop on each of the subsets.  The
improvement on F-score over iterations is steady and consistent; this trend is
made obvious in Figure~\ref{f:improvement}.  The found parameters seemed to
generalize well across subsets, in spite of the difference on orthographic
representation and usage\footnote{AS and CityU are in traditional Chinese; MSR
and PKU in simplified Chinese.}.  The overall performance is given in
Table~\ref{t:comparison2}.  Our incremental learning method achieved {in}
F-score above 80.0 on all four subsets, and it compares favorably with weaker
baselines such as MDL-based methods and DLG.  On two sets AS and PKU, incremental
learning outperformed the strong baseline ESA, winning out by 4.0 and 1.3 {in}
F-score respectively.  On the other two sets MSR and CityU, incremental
learning seems on a par with ESA: it improved over the baseline by 0.3 on MSR,
but it also lagged behind by 0.3 on CityU.  

To conclude, our result suggests that incremental learning is effective and
efficient for unsupervised word segmentation.  The main reason is that
penalized likelihood has better coverage in the hypothesis space; through
varying the parameters $\alpha$ and $\beta$, we are able to explore many more
search paths than an ordinary greedy method can.  Moreover, the information
criteria such as AIC and MDL allow us to make a guess fairly efficiently on the
solution quality.  It is the combination of the two that makes incremental
learning a plausible alternative for this specific application.

\section{Post-Hoc Analysis}\label{s:post-hoc-analysis}

\begin{figure*}[t]
  \centering 
  \begin{minipage}{0.30\textwidth}
    \includegraphics[width=\linewidth]{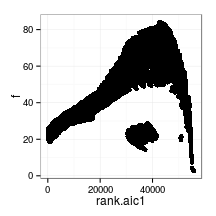}
  \end{minipage}
  \begin{minipage}{0.30\textwidth}
    \includegraphics[width=\linewidth]{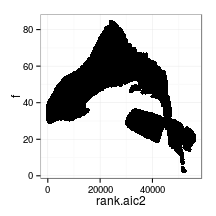}
  \end{minipage}
  \begin{minipage}{0.30\textwidth}
    \includegraphics[width=\linewidth]{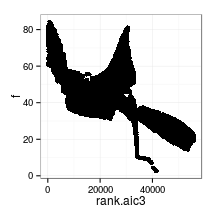}
  \end{minipage}
  \begin{minipage}{0.30\textwidth}
    \includegraphics[width=\linewidth]{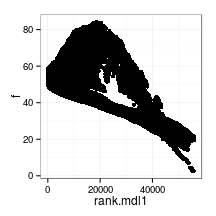}
  \end{minipage}
  \begin{minipage}{0.30\textwidth}
    \includegraphics[width=\linewidth]{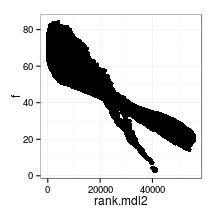}
  \end{minipage}
  \begin{minipage}{0.30\textwidth}
    \includegraphics[width=\linewidth]{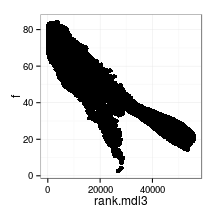}
  \end{minipage}
  \caption{F-score plotted against various information criteria on the full trace.}
  \label{f:interaction}
\end{figure*}

% \begin{figure*}
  % \centering 
  % \begin{minipage}{0.49\textwidth}
    % \includegraphics[width=\linewidth]{interaction-xlogx-top-both.pdf}
  % \end{minipage}
  % \begin{minipage}{0.49\textwidth}
    % \includegraphics[width=\linewidth]{interaction-x2-top-both.pdf}
  % \end{minipage}
  % \caption{This is a fake figure}
% \end{figure*}

We have initial evidence to support that information criteria can suggest
\emph{good} ways to segment words.  It nevertheless raises a question on the
reliability of these meta-heuristics.  By reliability, we mean how consistent
the prediction made matches the truth.  This has little to do with the exact
decision values; in our application, we care only about whether an information
criterion assigns sensible ranking to the given hypotheses.  In this section,
we discuss two ways to examine this relation.

\paragraph{Spectral Correlation}  The first one is to look for \emph{spectral
patterns} for the response values in the hypothesis space; a simple
visualization as in Figure~\ref{f:heatmap} would do the trick.  We plot the
F-score, \AIC{3}, and \MDL{2} values as individual heat maps on the parameter
search plane.  It is interesting to note that, on both penalties, the best
segmentation performance concentrates on a cone-like region with its peak
facing the y-axis.  For any $\alpha$, performance declines as $\beta$ moves
away from this region either towards zero or infinity.  As $\alpha$ increases,
the spread gets larger and results in more gentle decline, so it is not hard to
imagine the true performance contour as a mountain that has a fat end at large
$\alpha$.  We found that in general \AIC{3} and \MDL{2} assign similar rankings
to hypothesis, although the \AIC{3} ranking seems more fine-grained.  Both
criteria have missed on the true optimal solutions, but \MDL{2} does a bit
better in modeling by using two ``valleys'' to cap the true region.

\paragraph{Rank Correlation}  The second angle is to check on the rank
correlation.  We computed Spearman's $\rho$ between token F-measure and all the
information criteria used in the experiment.  For comparison, we also used
the full trace, which is the intermediate outputs collected every 100 iteration
for all the combinations.  The result is given in Table~\ref{t:correlation}.
We notice that on the output set both AIC and MDL give strong, negative
correlation, i.e., $\rho < -0.7$, with the true performance, with one
exception \MDL{3} that shows medium correlation.  This however does not match
our empirical result.  Note that the output set alone is not sufficient to draw
conclusion on the predictability over the entire hypothesis space; the criteria
may still assign good ranking to a bad hypothesis not covered in the output
set.

The rank correlation on the full trace has shown a better fit to the true
result.  Only 4 criteria, including \AIC{3}, \MDL{1}, \MDL{2}, and \MDL{3}, are
strongly correlated with the F-score.  It was suggested that MDL correlates
well with data under generative assumptions that vary in complexity.  The best
correlation is found on \MDL{2}, which sets record by achieving -0.96 on $x
\log x$ penalty.  This has once again been evidence in favor of our empirical
findings.  

The analysis can be further enhanced with some visual cues as given
in Figure~\ref{f:interaction}, which plots F-score against ranking for all six
decision criteria using the full trace data.  It suggests that \AIC{3},
\MDL{2}, and \MDL{3} are all able to find good segmentation hypotheses
reliably, although \AIC{3} has some serious problem in pushing solutions
towards mid-ranked regions.

\section{Conclusions}\label{s:conclusion}

In this paper, we introduce an incremental learning algorithm for unsupervised
word segmentation that combines probabilistic modeling and model selection in
one framework.  We show with extensive experiments that simple ideas, such as
the super-additive penalties and higher-order generative assumptions in model
selection, can achieve very competitive performance in both phonemic and
orthographic word segmentation.  Our algorithm is very efficient and scales
well on large corpora, making it more useful than other algorithms in
real-world applications.  Besides all that, this framework is general enough so
that it is easy to replace the objective function or model selection method
with more sophisticated ones.  For our future work, we will focus on enhancing
accuracy by incorporating more complex structural assumptions such as syntax
and context.

% In this paper we introduce a general incremental learning framework for
% unsupervised word segmentation, featuring a novel use of super-additive
% penalties to model the cognitive load produced by subword composition.  This
% design coupled with model selection using higher-order generative
% assumption has achieved very competitive performance against state of the art
% methods.  Our future work will focus on deploying this incremental
% learning method to other lexical acquisition tasks.

% \section*{Acknowledgements}

\bibliography{report}
\bibliographystyle{plain}

\end{document}